\documentclass[a4paper]{easychair}

\usepackage{subfigure}
\usepackage{graphics}

\usepackage{algorithm}
\usepackage{algorithmic}
\usepackage{graphicx}
\begin{document}

%% Front Matter
%%
% Regular title as in the article class.
%
\title{Mobile Phone Based Vehicle License Plate Recognition for Road Policing }

% \titlerunning{} has to be set to either the main title or its shorter
% version for the running heads. Use {\sf} for highliting your system
% name, application, or a tool.
%
\titlerunning{Mobile License Plate Recognition}

% For only the editors. Authors, please keep this commented out
%\volumeinfo
%	{G. Sutcliffe, A. Voronkov} % editors
%	{2}                         % number of editors
%	{{\easychair} 1.0, 2008}    % event
%	{1}                         % volume
%	{1}                         % issue
%	{1}                         % starting page number

% Authors are joined by \and and their affiliations are on the
% subsequent lines separated by \\ just like the article class
% allows.
%
\author{Lajish V. L.\\
TCS Innovation Labs - Mumbai\\
Yantra Park, Thane(W) - 400601\\
\url{lajish.vl@tcs.com}
\and
Sunil Kumar Kopparapu\\
TCS Innovation Labs - Mumbai\\
Yantra Park, Thane(W) - 400601\\
\url{sunilkumar.kopparapu@tcs.com}\\
%\and
%Andrei Voronkov\thanks{Masterminded EasyChair}\\
%University of Manchester\\
%Manchester, U.K.\\
%\url{andrei@voronkov.com}\\
}

% \authorrunning{} has to be set for the shorter version of the authors' names;
% otherwise a warning will be rendered in the running heads.
%
\authorrunning{Lajish, Kopparapu}

\maketitle

%------------------------------------------------------------------------------
% Abstract
%
\begin{abstract}

Identity of a vehicle is done through the vehicle license plate by traffic 
police in general. 
Automatic vehicle license plate recognition has several applications in intelligent traffic 
management systems. The security situation across the globe and particularly in India 
demands a need to equip the traffic police with a system that enables them to get instant details of 
a vehicle. The system should be easy to use, should be mobile, and work $24 \times 7$. 
In this paper, we 
describe a mobile phone based, client-server architected, license plate recognition system. 
While we use the state of the art image processing and pattern recognition algorithms tuned for 
Indian conditions to automatically recognize non-uniform license plates, the main
contribution is in creating an end to end usable solution. 
The client application runs on a  
mobile device and a server application, with access to vehicle information database, is hosted 
centrally. The solution enables capture of license plate image captured by the
phone camera and passes to the server; on the server 
the license plate number is recognized; the data associated with the number
plate is then  sent back to the mobile device, instantaneously. 
We describe the end to end system architecture in 
detail.  A working prototype of the proposed system has 
been implemented in the lab environment.

\end{abstract}

%------------------------------------------------------------------------------
\section{Introduction}
\label{sect:introduction}

License Plate Recognition (LPR) systems are usually designed to read vehicles
license plate and automatically recognize license plate number in ASCII.
of vehicles passing through a certain point. Systems are readily available for mass 
surveillance that utilizes optical character recognition (OCR) and hardware capable of reading 
license plates of moving vehicles\cite{web_apr,gonz,c009} but still continue to hold the
interest of researchers \cite{1573448,1553051,1512704,1525345,1426776,1444530,1157149,1172371,1138707} 
mainly because of several challenges
that exist.
Low quality images due to severe illumination conditions, vehicle motion,
viewpoint and distance changes, complex background are some of
challenges. LPR is preceded by localization of license plate. \cite{1172371} use
a feature-based
license plate localization algorithm that copes with multi-object problem in
different image capturing conditions which they claim is robust against
illumination, shadow, scale, rotation, and weather condition. 
Vahid et. al.  \cite{1553051} address some of these problems by
using
intensity variance and edge density
image enhancement methods. More recently Luis et al 
\cite{1573448}
have suggested the use of
artificial neural networks for license plate detection and Cristian et al
\cite{1512704}
suggest a mechanism to fuse decisions for improving the license plate
recognition.
Some systems make use of infrared cameras to increase the 
efficiency of the system. LPR systems can be used for vehicle identification
\cite{c008}, enforcement, collect 
electronic tolls\cite{toll}, traffic monitoring and travel management \cite{travel_time}. 
In all these systems the camera is 
fixed and is therefore only able to scan the vehicle passing through a particular point. 
On the other 
hand, mobile LPR system has become a necessity for law enforcement especially 
with increasing volumes of vehicles being added to Indian roads every year. 

Traffic police not only 
make sure that the vehicles on the road follow traffic rules but also make sure that the 
vehicles have necessary authorizations to be on the road. 
In any large city the traffic police 
inspectors can police a fraction of the city and when ascertaining details of
the vehicle have to rely solely on the 
details provided by the vehicle driver. 
In this paper we address the problem of equipping the traffic 
police with a mobile tool that can be used to get on the spot exact details of a vehicle on the road 
$24 \times 7$. We 
propose a mobile phone based vehicle license plate recognition system to assist the traffic police.
%\subsection{Need for License Plate Recognition on Mobile Phone}
%\label{sec:pot_applications}
The mobile phone based LPR has several applications.  They could be used to 
identify (a) vehicles involved in road accidents, (b) trace stolen vehicles, 
(c) trace VIP escorting vehicles, (d) monitor and issue 
memo for violation of traffic rules. 
The paper is organized as follows: in Section \ref{sect:sys_architecture} we 
describe the overall system architecture of the solution. Experimental setup
and  results are discussion of Section \ref{sect:results} and conclude in
Section \ref{sect:concl}.
\section{The Solution Architecture}
\label{sect:sys_architecture}

\begin{figure}
	\begin{centering}
	\includegraphics[height=0.5\textwidth]{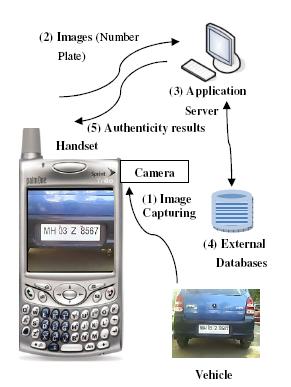}
	\caption{The high-level client-server architecture of the solution.}
	\label{fig:high_level_architeture}
	\end{centering}
\end{figure}

The proposed solution has two parts (a) mobile client which enables capture of
the license plate of a vehicle and send the image to a remote server and (b) a
remote server which has a LPR software and access to an external database which
has vehicle number and vehicle information association details.
Figure \ref{fig:high_level_architeture} shows the high-level client-server solution architecture. 

%\begin{figure}[htb!]

%------------------------------------------------------------------------------
%\subsection{Client Functionalities }
%\label{sect:client_funct}

\begin{figure}
\subfigure[Client Side]{
	\includegraphics{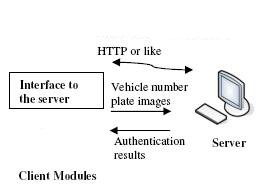}
	\label{fig:client_m}
}
\subfigure[Server Side]{
	\includegraphics{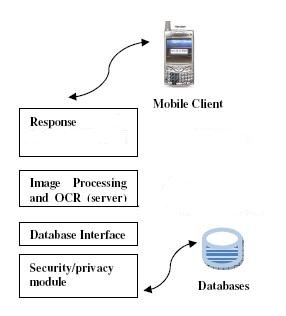}
	\label{fig:server_m}
}
\caption{Solution Architecture Details}
\end{figure}

The major client side components (Figure \ref{fig:client_m}) are  
(a) a server interface module and (b)
a media interface (capture image) module.
The client module communicates with the server using the HTTP. 
In a typical working scenario, the client application requests the user  
to capture the image of the vehicle license plate
and enables the camera capture mode on the mobile phone. 
Once the image is captured, the image is pushed to the server for recognition.  
On receiving information from the server, the  client software displays the 
obtained information about the vehicle 
as a text message to the user. Figure \ref{fig:client_m} shows the client side solution 
architectures in detail. 

Note that the client application can be downloaded over the air (current implementation is on BREW but 
can be enabled for multiple mobile devices and multiple OS).

%\begin{figure}[htp]
%	\begin{centering}
%	\includegraphics{client_m}
%	\caption{Solution Architecture in greater detail (Client side)}
%	\label{fig:client_m}
%	\end{centering}
%\end{figure}

%------------------------------------------------------------------------------
%\subsection{Server Functionalities }
%\label{sect:server_funct}

The major components on the server (Figure \ref{fig:server_m}) include (a) image processing and pattern recognition modules for 
recognizing license plate and (b) a database interface.
% A regional language interface module is also 
%present in the server end which provides the vehicle information, ownership details or complaint 
%registration feedback in regional languages to the end user. Figure~\ref{fig:server_m} shows the 
%server side solution architecture in greater detail.
%\begin{figure}[htp]
%	\begin{centering}
%	\includegraphics{server_m}
%	\caption{Solution Architecture in greater detail (Server side)}
%	\label{fig:server_m}
%	\end{centering}
%\end{figure}
 The server can interact with external databases (for example, in case of
stolen vehicles \cite{web_sv}) 
and find the association between vehicle number and its status against a stolen
vehicle. The ownership details (name, contact address, number), vehicle details (make, model, engine number 
etc.), tax details and previous complaints registered against the vehicle can
also be obtained if a database is available.
%This collective 
%information can be used with better effect by the traffic police especially under suspicious 
%situations.\\ 
The server works on the image sent by the client while maintaining a link between the 
image uploaded and the mobile device from which the image was uploaded 
through a unique session number. A module to recognize vehicle 
registration number from the image forms the heart of this solution and is
described in greater detail in Section \ref{sect:np_recog_eng}.
% describes the 
%functionalities of the license plate recognition engine in detail.

\section{License Plate Recognition}
\label{sect:np_recog_eng}

\begin{figure}[htb!]
	\begin{centering}
	\includegraphics{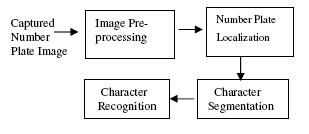}
	\caption{Block diagram of LPR engine}
	\label{fig:re_block_diangram}
	\end{centering}
\end{figure}

The LPR engine resides on the server and is a hub of several image processing and 
pattern recognition modules. The important functional units include image
pre-processing, license 
plate localization, character segmentation and character recognition \cite{delta,np_aus}. 
Figure \ref{fig:re_block_diangram} shows the block diagram of the LPR engine.
After initial pre-processing of the LP image, the next task is that of locating
the exact position of the LP in the image (LP localization). Once the LP is
located the actual LPR happens.

\begin{figure}
\begin{centering}
\subfigure[]{\includegraphics[width=0.2\textwidth]{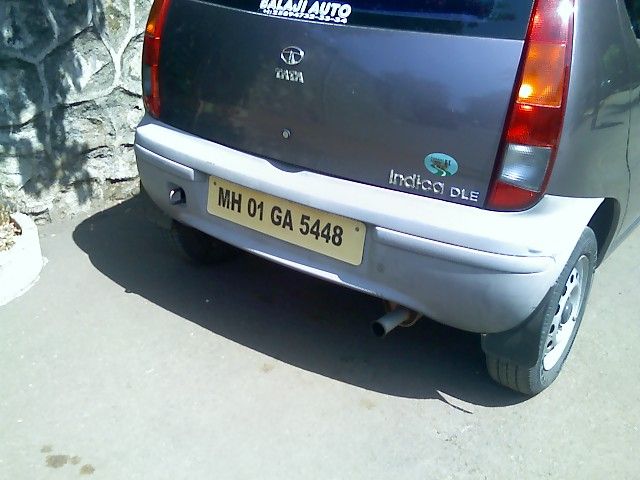}}
\subfigure[]{\includegraphics[width=0.2\textwidth]{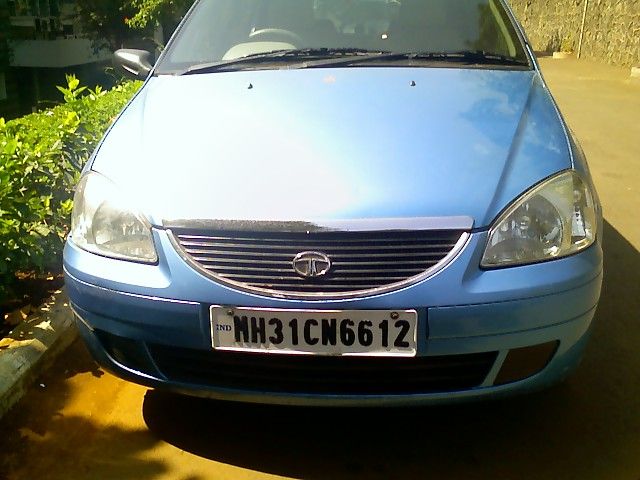}}
\subfigure[]{\includegraphics[width=0.2\textwidth]{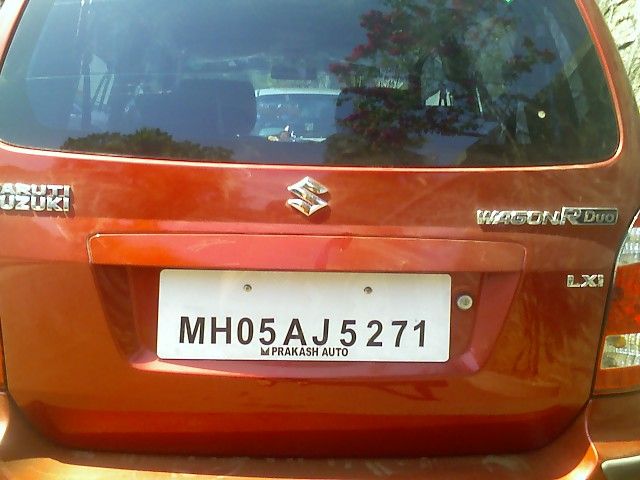}}
\subfigure[]{\includegraphics[width=0.2\textwidth]{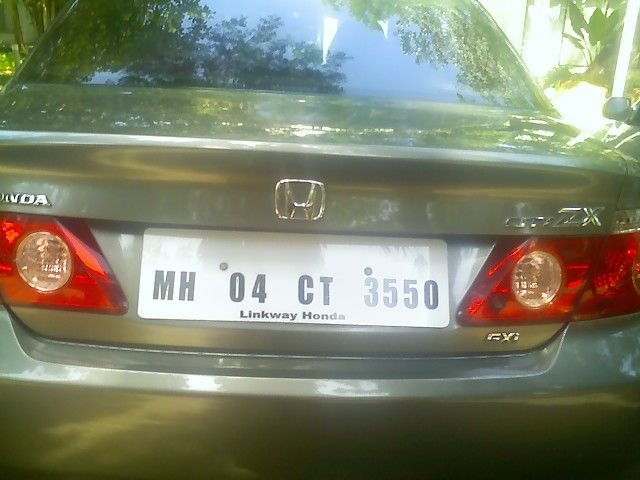}}

\subfigure[]{\includegraphics[width=0.2\textwidth]{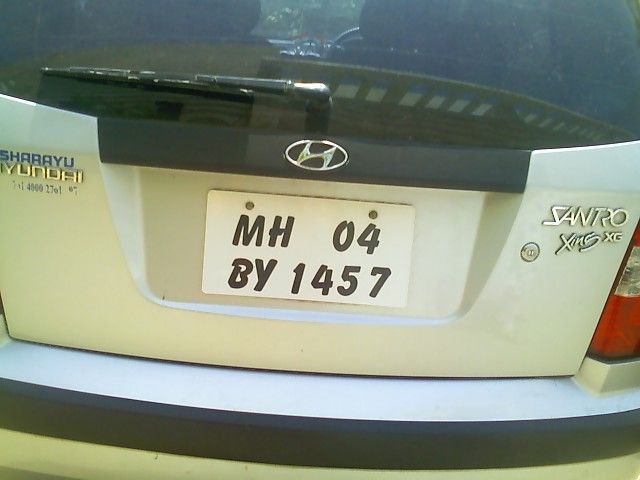}}
\subfigure[]{\includegraphics[width=0.2\textwidth]{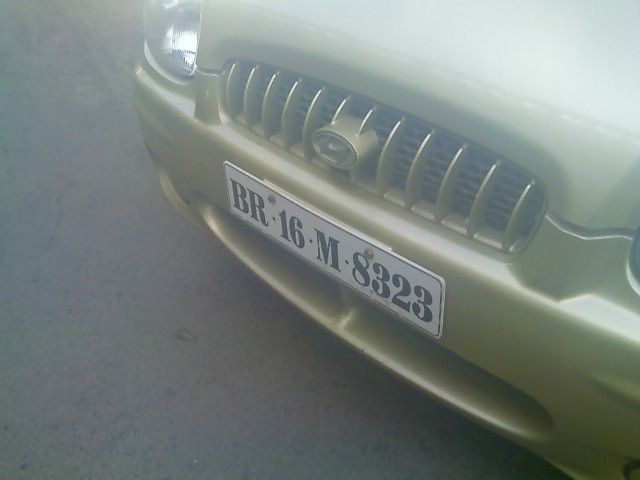}}
\subfigure[]{\includegraphics[width=0.2\textwidth]{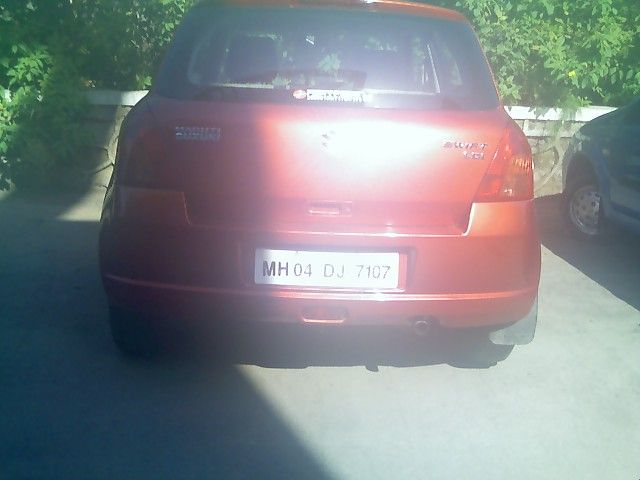}}
\subfigure[]{\includegraphics[width=0.2\textwidth]{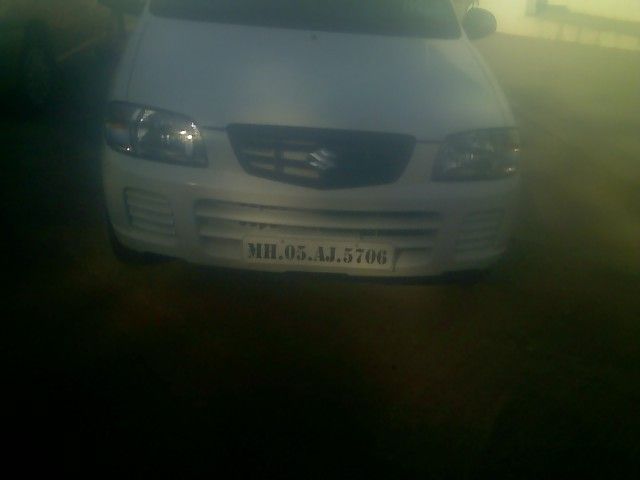}}
\caption{Images of the license plates taken using mobile phone cameras}
\label{fig:np_samples}
\end{centering}
\end{figure}

Actual sample images of vehicle license plates captured by mobile phone cameras are shown in 
Figure \ref{fig:np_samples}. It can be observed that the input images have a great deal of 
variability 
%{\bf [Lajish: Describe the variability]} 
in terms of variability in lighting conditions, shadows on the plate region,
plates from front or rear part of the vehicle with messages or icons, skewed images, background and foreground of the number plate and one line and 
two line writing, the recognition system needs to work under all these conditions.

%\subsubsection{LP Pre-processing}
%\label{sect:Img_processing}

\noindent {\bf License Plate Pre-processing:}
The color image captured from the camera is first converted into a gray scale
image and is  then binarized using Otsus global thresholding technique
\cite{otsu1}. Note that we could use an adaptive 
threshold technique to get a binary image but our analysis with a range of LP
images showed that Otsus thresholding technique works well. 
We also use, skew correction up to about $\pm 5$ degrees and slant normalization  
before license plate localization is carried out. 

\begin{figure}
\begin{center}
\subfigure[Input]
{\includegraphics[width=0.35\textwidth]{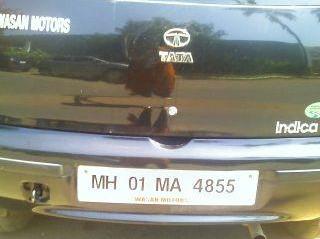}}
\subfigure[Output]
{\includegraphics[width=0.35\textwidth]{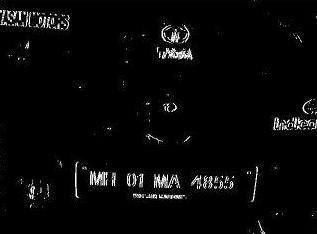}}
\caption{Sample vertical edge output}
\label{fig:v_edge}
\end{center}
\end{figure}

\noindent {\bf License Plate Localization:}
Determining 
the location of the LP within the captured image automatically is a critical task for 
successful recognition. The aim is to obtain a region of interest (usually a rectangular window) in 
the image (say $I(x,y)$ of size $M \times N$) that includes the license plate of the vehicle.  
For this purpose, the edge map 
$E(x,y)$ of $I(x,y)$ is initially computed by applying the Sobel 
edge operator in the vertical direction since presence of characters contribute more
edges in vertical direction (see Figure \ref{fig:v_edge}). Note that the edge values are real numbers.    %and the horizontal directions. 
We eliminate all but the top 3\% of the edge 
intensity pixels and obtain a reduced binary edge map $E(x,y)$. 
We compute variance ($\sigma^2_E(x)$) of the each row of the edge image
($E(x,y)$) and calculate the maximum variance ($\sigma_{max}^2$). We select 
all the rows ($x$) with  $\sigma^2_E(x) \ge 0.5 \times \sigma_{max}^2$.
We mark and collect contiguous rows as regions (say $\xi_1, \xi_2 .... $); this
isolates the license plate in the horizontal direction.
%$E(x,y)$ is broken into overlapping sub-images and a coarse search for the license plate 
%region is carried out using edge density information computed over the
%sub-region of the image $E(x,y)$.  For a sub-region $r$ with the left-top corner at $(x_1,y_1)$ and the right bottom corner
%at $(x_2,y_2)$ the edge density feature is defined as,
%%\begin{displaymath} 
%\[ f_r = \frac{1}{a_{r}} \sum_{x=x_1}^{x_{2}} \sum_{y=y_1}^{y_{2}} E(x,y) \]
%%\end{displaymath}
%where $a_{r}$ is the area of the sub region and is defined as, $a_{r} = (x_{2} - x_{1} 
%+1) (y_{2} -y_{1}+1)$. Higher $f$ is indicative of the existence license plate
%region. This is primarily due to the fact that LP are generally high contrast
%regions and the presence of characters contributes to more edges an higher $f$. 
%{\bf [Lajish: Not clear if we are doing this after finding $f$?]} 
%We first search vertically downwards considering 
%overlapping horizontal strips of size 30 x N with a $50 \%$ overlap. Horizontal strip size is 
%heuristically determined based on the approximate size of a license plate. 
%The top three 
%high density regions are ear marked as candidate regions for the presence of
%license plate. 
%To identify the actual 
%region, we examine for uniformly spaced valleys {\bf [Lajish: More
%Explanation - this is very sketchy. We should write what we are doing.]} in the vertical profile in each of the candidate 
%regions. 
To perform vertical cropping of the selected regions, we examine the edge strength 
(by observing the vertical profile of the horizontally segmented 
region) in vertical direction ($E(x,y)$ in the $y$ direction). 
The columns with 50\% of the maximum edge strength are retained and the co-ordinates of the license 
plate region are identified. We end up with one or more candidate license plate regions.

\noindent{\bf Character Segmentation:}
%\label{sect:char_seg}
After identifying the regions of interest the next step is to segment the
characters from the LP  as a spatial sequence of alpha-numeric number 
(which gives the exact number of the vehicle). 
Segmentation of the characters 
written in different formats (single line, double line, or in different
orientation) is a challenge. 
The character segmentation (isolation) process involves use of connected
component analysis to segment the characters \cite{c010}. Post processing on
the identified connected components is performed to eliminate non-character
like logos. Post processing involves use of spatial position of the connected
components (all the characters have more or less the same $x$ coordinates but
varying $y$ coordinate), the aspect ratio of the bound box on the connected
component (all the characters are more or less of the same size) and
also use of high level information like the maximum number of characters
expected in the license plate, etc.
%Algorithm \ref{algo:char_seg} describes the charcater segmentaion process.
%\begin{algorithm}
%\begin{algorithmic}[1]
%\STATE {Input image is the resultant Image after Plate Localization} 
%\STATE {Apply Pre-processing on the localized image to obtain a binary image (RGB to Gray, Median filtering and Global thresholding)}
%\STATE {Identify and label connected components ($CC$) with position ($p$)} 
%\STATE {Find total number of CC (say, $n$)}
%       {Find size ($z$) and Aspect Ratio (AR) for each component}
%       {Let maximum number of characters expected in an Indian number plate $= N$}
%\STATE {If If $n \ge N$ 
%   Eliminate minimum ($n - E$) CCs based on the decision logic and thresholds fixed on $z$ and AR 
%   else  
%   Perform segmentation in case the selected CC represents two characters (based on aspect ratio) 
%   followed by size normalization}
%\STATE {The selected CCs represent license plate characters. Find the exact sequence of characters 
%   based on the position ($p$) of each component} 
%\end{algorithmic}
%\caption{Steps in character segmentation}
%\label{algo:char_seg}
%\end{algorithm}
The isolated characters are then size normalized using affine transformation and bilinear interpolation techniques 
\cite{size_nor}.

%{\bf [Lajish: What about challenges when two characters are joined and you see
%them as a single component etc .....]}
% This helps in comparing the characters with the template (set of all characters and 
%digits used to constitute vehicle registration numbers in India) which are stored in a specified 
%size.

\noindent {\bf Character Recognition:}
%\label{sect:char_recog}
The segmented characters are recognized using template matching technique. 
Here the segmented character is compared with all the 
reference character set and the similarity is computed  (correlation
coefficient).  
The reference character with the highest correlation is chosen as the
identified character.
%The mathematical formulation of the method is explained below.
%Let $ F_{x}(j, k)$ and $F_{r}(j, k)$ for $1 \le j \le J$ and $1 \le k \le K$ represent 
%the image to be recognized and 
%the reference template respectively. 
Note that the template database consists of all English capital 
letters and Arabic numerals.
%The normalized cross-correlation between the image pair is defined as follows.
%%{\bf [Lajish: How do we get a single number from this $M \times N$ $R(m,n$?]}
%\[ 
%R(m,n) = \frac{\sum_{j} \sum_{k} F_{x} (j,k) F_{r} \left (j-m +
%\frac{(M+1)}{2}, k - n + \frac{(N+1)}{2} \right )}  
%{\sqrt{\sum_{j}\sum_{k} |F_{x} (j,k)|^{2}}
%\sqrt{\sum_{j}\sum_{k} \left | F_{r} \left (j- m + \frac{(M+1)}{2}, k - n +
%\frac{(N+1)}{2} \right )\right |^{2}}}
%\] 
%from $m = 1, 2 \cdots, M$ and $n = 1, 2, \cdots, N$.
%
%To find the highest correlated character we find the sum of the diagnonals of
%$R$ ($R(m,m)$) for each image pair and identify the
%test image which is having maximum value for it.
%
%The most confused characters found in the set are (B, 
%8), (D, O, 0), (E, F), (S, 5), (Z, 2). 
%The performance of the recognition engine is aided by 
%incorporating the prior information about the format of the Indian license plates like set of state 
%and district codes and its combinations used to form the registration number.
% Another important 
%module on the client is the location identification module. This module helps to locate the location 
%of the mobile device. This module is designed as a Location Based Services (LBS) platform based on 
%Global Positioning System (GPS). The location information captured here is sent to the server along 
%with the image of the vehicle license plate. The location information captured along with the number 
%plate information of the vehicle present at a specific location helps in vehicle tracking 
%\cite{travel_time}.
Algorithm \ref{algo:npr} describes the proposed solution.
\begin{algorithm}
\begin{algorithmic}[1]
\STATE{Convert Colour ($I_c(x,y)$) image to Grayscale  ($I(x,y)$) }
\STATE{Extract vertical edge map using Sobel edge detection algorithm
($E_v(x,y)$)}
%\STATE{Apply thresholding on edge image to eliminate weak edges }
\STATE{Compute variance ($\sigma^2_E(x)$) of the each row of the edge image and compute the
maximum variance ($\sigma_{max}^2$})
\STATE{Select all the rows ($x$) with  $\sigma^2_E(x) \ge 0.5 \times \sigma_{max}^2$}
\STATE{Mark and collect contiguous rows as regions (say $\xi_1, \xi_2 .... $)}
%\STATE{Perform rule based elimination of surplus regions}
\STATE{Binarise all the selected region ($I_{\xi_1}, I_{\xi_2}, \cdots $) using Otsu's algorithm}
\STATE{Perform vertical cropping of the identified region based on vertical
profile}
\STATE{Apply morphological dilation for bridging the discontinuities within
the character segments in the plate region.}
\STATE{Perform character segmentation based on connected components }
\STATE{	Perform rule (area /aspect ratio / existence ratio) based
elimination of the non character components.}
%{\bf [Lajish: We need to use this informstion in the actaul text.. I think we
%are describing something else in the prose and something else in the algorithm
%here??]}
%\STATE{
%1.	Colour to Grayscale conversion 
%2.	Perform Sobel edge detection to get vertical edges 
%3.	Apply thresholding on edge image to eliminate weak edges 
%4.	Compute variance of the each row of the edge image
%5.	Select the rows with variance greater than $50 \%$ of the maximum variance
%6.	Identify the regions with contiguous rows
%7.	Perform rule based elimination of surplus regions 
%8.	Binarise the selected region using Otsu's algorithm
%9.	Perform vertical cropping of the identified region based on vertical profile 
%10.	Apply morphological dilation for bridging the discontinuities within the character segments in the plate region.
%11.	Performed character segmentation based on connected components 
%12.	Performed rule (area /aspect ratio / existence ration) based elimination of the non character components.
%13.	Correlation based template matching is used for character segmentation.
%}
\end{algorithmic}
\caption{Steps in LPR}
\label{algo:npr}
\end{algorithm}

\section{Experimental Results}
\label{sect:results}

%{\bf [Lajish: You will need to give details of the database that we collected
%in terms of their variability and number also]}

There is no standard test bed, to the best of our knowledge,
to test the performance of a LPR system; hence we collected our own data.
Additionally, we needed to work on mobile camera resolution images. We
collected images as follows: (a) the images  are colour images of actual Indian license plates taken under various
conditions, (b) all images are taken in $640 \times 480$ resolution or lower  
with different mobile phones at our disposal. The database contained a total of
$871$ images of $230$ different
vehicles. 
The results of image processing algorithms 
used in the recognition including plate localization, binarization, character segmentation and size 
normalization are given below. 

Sample experimental outputs shown in Figure \ref{fig:results}. Results show that the 
license plate recognition engine is capable of handling images which are skewed
and also able to handle when license plate characters come in more than one
row.
%{\bf [Lajish: Need actual recognition results if available ... else we need to
%discuss what exactly we need to do.]}
Experiments were performed to measure the accuracy of LP region of interest
extraction and the LP recognition. The complete results will be presented in
the final version of the paper.

\begin{figure}
\subfigure[]{
\includegraphics[width=0.45\textwidth]{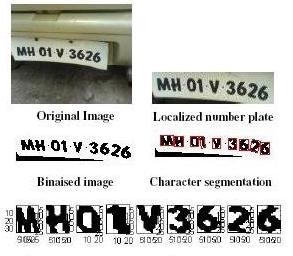}
%\caption{Experimental Results-1}
%\label{fig:re_result1}
}
\subfigure[]{
\includegraphics[width=0.45\textwidth]{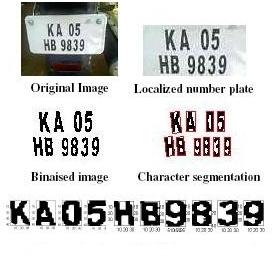}
%\caption{Experimental Results-1}
%\label{fig:re_result1}
}
\caption{Some Experimental Results}
\label{fig:results}
\end{figure}

\section{Conclusions}
\label{sect:concl}

In this paper, we presented the solution architecture and a sequel of algorithms for the recognition of 
LP of a vehicle in Indian context using mobile phone camera images. 
This system has been tested over a large number of real images with very
encouraging results. The proposed system is fast and yields 
robust recognition results 
%and provides a mechanism to communicate with the user even in regional 
%language. 
%Solution mobility is the additional advantage of this system compared to the surveillance camera 
%based existing systems. 
Almost all existing vehicle license plate recognition systems use one or more 
fixed cameras which put a restriction on mobility. 
In contrast, the system proposed uses mobile phone camera to capture 
license plate images, meaning we have a vehicle plate recognition system which is mobile (we can take 
it to any location). Mobility facilitates handling vehicles parked anywhere 
or when vehicles are stopped for inspection by the traffic police. We believe, mobile phone based 
systems are currently not available anywhere in the world and especially for Indian license plates. 
Comparison with any existing LPR system is difficult because of differences in
the working environment of the proposed system with a existing LPR system. 
%(a) the system can not actually be used in a moving vehicle scenario because it 
%is not very convenient to capture the license plate of the vehicle without coming in front or back of 
%the vehicle without disrupting the traffic and putting the person capturing the image in danger.
As a future enhancement, location information can be captured along with the 
LP images and can be used in vehicle tracking application. The average speed and travel 
time between two points can be calculated and presented in order to monitor traffic flow and load in 
a specific area of a city. 
%This system can also be redesigned for multinational vehicle license 
%plates in future. 

%\label{sect:bib}
%\bibliographystyle{plain}
%\bibliographystyle{alpha}
\bibliographystyle{unsrt}
\bibliography{easychair}

\begin{thebibliography}{10}

\bibitem{web_apr}
Adaptive~Recognition Hungary.
\newblock http://www.anpr.net/.

\bibitem{gonz}
Gonzalez R.C and Herrera J.A.
\newblock Apparatus for reading a license plate.
\newblock Patent US4817166, 1989.

\bibitem{c009}
Tyan J.-K. and Neubauer C.
\newblock Character segmentation method for vehicle license plate recognition.
\newblock Patent EP1085456, 2001.

\bibitem{1573448}
Luis Carrera, Marco Mora, Jos\'{e} Gonzalez, and Francisco Aravena.
\newblock License plate detection using neural networks.
\newblock In {\em IWANN '09: Proceedings of the 10th International
  Work-Conference on Artificial Neural Networks}, pages 1248--1255, Berlin,
  Heidelberg, 2009. Springer-Verlag.

\bibitem{1553051}
Vahid Abolghasemi and Alireza Ahmadyfard.
\newblock An edge-based color-aided method for license plate detection.
\newblock {\em Image Vision Comput.}, 27(8):1134--1142, 2009.

\bibitem{1512704}
Cristian Molder, Mircea Boscoianu, Iulian~C. Vizitiu, and Mihai~I. Stanciu.
\newblock Decision fusion for improved automatic license plate recognition.
\newblock {\em WSEAS Trans. Info. Sci. and App.}, 6(2):291--300, 2009.

\bibitem{1525345}
R.~Huang, H.~Tawfik, and A.~K. Nagar.
\newblock Licence plate character recognition based on support vector machines
  with clonal selection and fish swarm algorithms.
\newblock In {\em UKSIM '09: Proceedings of the UKSim 2009: 11th International
  Conference on Computer Modelling and Simulation}, pages 101--106, Washington,
  DC, USA, 2009. IEEE Computer Society.

\bibitem{1426776}
Rentian Huang, Hissam Tawfik, and Atulya Nagar.
\newblock Licence plate character recognition using artificial immune
  technique.
\newblock In {\em ICCS '08: Proceedings of the 8th international conference on
  Computational Science, Part I}, pages 823--832, Berlin, Heidelberg, 2008.
  Springer-Verlag.

\bibitem{1444530}
Hua Xu and Zheng Ma.
\newblock A practical design of gabor filter applied to licence plate character
  recognition.
\newblock In {\em ICCSIT '08: Proceedings of the 2008 International Conference
  on Computer Science and Information Technology}, pages 397--401, Washington,
  DC, USA, 2008. IEEE Computer Society.

\bibitem{1157149}
Gang Li, Ruili Zeng, and Ling Lin.
\newblock Research on vehicle license plate location based on neural networks.
\newblock In {\em ICICIC '06: Proceedings of the First International Conference
  on Innovative Computing, Information and Control}, pages 174--177,
  Washington, DC, USA, 2006. IEEE Computer Society.

\bibitem{1172371}
Hamid Mahini, Shohreh Kasaei, Faezeh Dorri, and Fatemeh Dorri.
\newblock An efficient features - based license plate localization method.
\newblock In {\em ICPR '06: Proceedings of the 18th International Conference on
  Pattern Recognition}, pages 841--844, Washington, DC, USA, 2006. IEEE
  Computer Society.

\bibitem{1138707}
Danian Zheng, Yannan Zhao, and Jiaxin Wang.
\newblock An efficient method of license plate location.
\newblock {\em Pattern Recogn. Lett.}, 26(15):2431--2438, 2005.

\bibitem{c008}
Serkan Ozbay and Ergun Ercelebi.
\newblock Automatic vehicle identification by plate recognition.
\newblock {\em Proc. of World Academy of Science, Engineering and Technology},
  2005.

\bibitem{toll}
Davies P., Emmott N., and Ayland N.
\newblock License plate recognition technology for toll violation enforcement.
\newblock {\em Proceedings of IEE Colloquium on Image Analysis for Transport
  Applications}, 7:1--5, 1990.

\bibitem{travel_time}
Kanayama K., Fujikawa Y., Fujimoto K., and Horino M.
\newblock Development of vehicle-license number recognition system using
  real-time image processing and its application to travel-time management.
\newblock {\em Proceedings 41st IEEE Vehicular Technology Conference}, pages
  798--804, 1993.

\bibitem{web_sv}
Stolen Vehicles.
\newblock http://www.stolen.in/stolenvehicle.php.

\bibitem{delta}
Bailey D.G, Irecki D, Lim B.K, and Yang L.
\newblock Test bed for number plate recognition applications.
\newblock {\em Proceedings of First IEEE International Workshop on Electronic
  Design, Test and Applications}, 2002.

\bibitem{np_aus}
Leonard~G. C, Hamey, and Colin Priest.
\newblock Automatic number plate recognition for australian conditions.
\newblock {\em Proc. of Digital Image Computing: Techniques and Applications},
  2005.

\bibitem{otsu1}
Otsu.N.
\newblock A threshold selection method from gray-level histograms.
\newblock {\em IEEE Trans. Systems Man and Cybernetics}, 9:62--66, 1979.

\bibitem{c010}
Yang W and Prabir B.
\newblock Using connected components to guide image understanding and
  segmentation.
\newblock {\em Proceedings of IEE Colloquium on Image Analysis for Transport
  Applications}, 12:163--186, 2003.

\bibitem{size_nor}
Oliveira de~J.J.Jr. and Veloso.L.R.Jr.
\newblock Interpolation /extrapolation scheme applied to size normalization of
  character images.
\newblock {\em Proc. 15th Int. Conf. of Pattern Recognition, Barccelona,
  Spain}, pages 577--580, 2000.

\end{thebibliography}

\end{document}